\documentclass[11pt]{article}

\usepackage[final]{acl}

\usepackage{times}
\usepackage{latexsym}

\usepackage[T1]{fontenc}

\usepackage[utf8]{inputenc}

\usepackage{microtype}

\usepackage{inconsolata}

\usepackage{graphicx}

\usepackage{multirow}
\usepackage{comment}
\usepackage{booktabs}
\usepackage[normalem]{ulem}
\usepackage[xcolor,framemethod=tikz]{mdframed}
\usepackage{collcell}
\usepackage{hhline}
\usepackage{colortbl}
\usepackage{lingmacros}
\usepackage{subcaption}
\usepackage{pgfplots}

\usepackage{tikz}

\usepackage{amssymb}
\usepackage{amsmath}
\usetikzlibrary{positioning, calc, arrows.meta, backgrounds}
\usepackage{algorithm}
\usepackage{algpseudocode}
\usepackage{array}    
\usepackage{tabularx}
\usepackage{enumitem}
\usepackage{fontawesome5}

\usepackage{xcolor}
\usepackage{pgffor} 

\newcounter{wordcounter}
\newcommand{\coloredbreak}[3]{
    \setcounter{wordcounter}{0}
    \foreach \word in {#3} {
        \textcolor{#2}{\word}
        \stepcounter{wordcounter}
        \ifnum\value{wordcounter}=#1
            \\ \setcounter{wordcounter}{0}
        \else
            \ 
        \fi
    }
}

\usepackage[dvipsnames]{xcolor}

\pgfplotsset{compat=1.18} 
\newcommand{\skel}[1]{\textcolor{RoyalBlue}{\textbf{#1}}}  
\newcommand{\depA}[1]{\textcolor{BrickRed}{\textbf{#1}}}   

\newcommand{\head}[1]{\textcolor{Mulberry}{\textbf{#1}}}

\title{Language Acquisition Device in Large Language Models}

\author{Masato Mita \quad Taiga Someya \quad Ryo Yoshida \quad Yohei Oseki \\
  The University of Tokyo \\
  \texttt{\{mita, tsomeya, yoshiryo0617, oseki\}@g.ecc.u-tokyo.ac.jp}}

\begin{document}
\maketitle

\begin{abstract}
Large Language Models (LLMs) remain substantially less data-efficient than humans.
Pre-pretraining (PPT) on synthetic languages has been proposed to close this gap, with prior work emphasizing highly expressive formal languages such as $k$-Shuffle Dyck.
Inspired by the \emph{Language Acquisition Device (LAD)} hypothesis, which posits that innate constraints preemptively restrict the learner's hypothesis space to natural-language-like structure, we propose \emph{LAD-inspired PPT}: pre-pretraining on \textsc{MP-Struct}, a formal language whose strings encode hierarchical composition, feature-based dependencies, and long-distance displacement via \textsc{Merge}, \textsc{Agree}, and \textsc{Move}.
A brief 500-step PPT with \textsc{MP-Struct} matches strong formal-language baselines in token efficiency while additionally imparting a human-like resistance to structurally implausible languages (e.g., \textsc{Reverse}).
Analyzing simplified variants, we find that \textsc{MP-Struct Core} outperforms $k$-Shuffle Dyck despite not being definable in C-RASP (a formal bound on transformer expressivity), challenging the prior hypothesis that effective PPT languages must be both hierarchically expressive and circuit-theoretically learnable.
We show that \emph{functional landmarks}, which reduce dependency resolution ambiguity, are a key driver, suggesting that effective PPT design depends not only on expressivity but also on the accessibility of dependency resolution.

\end{abstract}

\begin{center}
    \faGithub~\url{https://github.com/osekilab/LAD-PPT}
\end{center}

\section{Introduction}

Large language models (LLMs) exhibit general linguistic abilities comparable to those of humans; however, their efficiency in language acquisition remains far inferior.
While humans acquire language from limited text, LLMs typically require orders of magnitude more data to achieve strong performance \cite{warstadt-etal-2023-findings}.
This gap suggests that current LLMs rely on learning from an overly permissive hypothesis space \cite{yun2020are}, leaving substantial room for improving learning efficiency through better inductive biases.

Recent work by \citet{hu-etal-2025-circuits} explores \emph{pre-pretraining (PPT)}, where models are first trained on synthetic sequences to acquire useful structural biases before standard pretraining.
They show that highly expressive formal languages, such as $k$-Shuffle Dyck, can improve token efficiency by exercising the model's ability to process hierarchical dependencies.
They further interpret these results through the \emph{expressivity hypothesis}, which posits that a formal language conferring a helpful inductive bias should be hierarchically structured in the sense of the \emph{Chomsky hierarchy~\cite{chomsky-hierarchy}} (either context-free or context-sensitive) and definable in C-RASP~\cite{yang2024counting}, a formal measure of \emph{circuit complexity} for transformers.

However, such formal languages prioritize abstract structural complexity and often lack key properties characteristic of natural language.
This raises the question: \textit{beyond Chomsky-hierarchy complexity and circuit complexity, do natural-language-like properties, such as dependencies conditioned on fixed hierarchical structure, also contribute to effective PPT?}
In this work, we approach this question from a complementary perspective.
Taking inspiration from the \emph{Language Acquisition Device (LAD)} hypothesis \cite{chomsky1965}, which suggests that innate structural constraints can restrict the hypothesis space and favor natural-language-like structure, we ask whether incorporating such constraints into synthetic sequences can further improve learning efficiency.
Guided by this idea, we design \textsc{MP-Struct}, a synthetic generator taking cues from the \emph{Minimalist Program (MP)}, which produces sequences where dependencies are embedded within a fixed hierarchical structure rather than freely interleaved.

We evaluate LAD-inspired PPT on \texttt{Pythia-1B}~\cite{biderman2023pythia}.
A brief 500-step PPT phase with \textsc{MP-Struct} consistently improves token efficiency over training from scratch and achieves performance comparable to strong baselines based on formal languages such as $k$-Shuffle Dyck.
Moreover, the resulting models exhibit a directionally asymmetric inductive bias: compared to $k$-Shuffle Dyck, \textsc{MP-Struct} shows greater resistance to directionally reversed sequences, consistent with the LAD-inspired design goal of favoring natural-language-like directional structure.

To better understand these effects, we analyze simplified variants of the generator.
Notably, \textsc{MP-Struct Core}, an idealized abstraction of our generator, achieves higher efficiency than $k$-Shuffle Dyck despite not being definable in C-RASP, a formal lower bound on the expressivity of future-masked soft attention transformers.
This observation is not fully explained by the expressivity hypothesis, which predicts that effective PPT languages should be hierarchically structured and definable in C-RASP.
Our analysis instead points to a complementary factor: the \emph{accessibility} of dependency retrieval.
While structures that encode dependencies purely through bracketed hierarchy (e.g., $k$-Shuffle Dyck) define valid dependencies, they may leave multiple plausible antecedents, potentially leading to higher retrieval ambiguity.
In contrast, \textsc{MP-Struct} and \textsc{MP-Struct Core} introduce explicit structural cues—\emph{functional landmarks}—that can make dependencies easier to locate, thereby reducing the effective search cost for attention-based models.
We hypothesize that these differences in accessibility contribute to the observed efficiency gains.
More broadly, this suggests that effective PPT design may depend not only on formal expressivity, but also on how structural information is organized to support efficient dependency retrieval.

\section{Related Work}

\subsection{LAD/UG and Minimalism}
A longstanding challenge in language acquisition is explaining how children converge on rich grammatical competence from comparatively limited and noisy input (often termed the \textit{poverty of the stimulus})~\cite{chomsky1965, Clark-etal-2011}.
The LAD hypothesis addresses this gap by proposing that learners are endowed with Universal Grammar (UG), a species-specific set of constraints that sharply restricts the hypothesis space of possible grammars.
From this perspective, UG serves as a strong innate inductive bias, filtering out ``non-human-like'' hypotheses \emph{a priori}.

In contemporary generative grammar, the \emph{Minimalist Program} (MP) refines this LAD/UG model by seeking to minimize the computational machinery of language to a small set of operations~\cite{MP,MI,DbP}.
A central component of this framework is \textsc{Merge}, a combinatory operation that builds hierarchical structure.
In addition, operations such as \textsc{Move} (displacement) and \textsc{Agree} (feature valuation) are commonly assumed to play roles in dependency formation and morphosyntactic licensing~\cite{MI,DbP}.

\subsection{Pre-pretraining on Synthetic Structures}

\label{sec:ppt_background}

\begin{algorithm}[t]
\caption{Data Generation Procedure (\textsc{MP-Struct})}
\label{alg:mg_gen}
\centering
\footnotesize

\hrule \vspace{2pt}

\begin{minipage}{0.98\linewidth}
\textbf{Notation:} $\mathcal{L}$: lexicon, $V/N/D$: lexical categories, $vP$: verb phrase, $T/C$: functional heads, $t$: trace, $u/iNum$: number features, $wh \in \{+,-\}$.
\end{minipage}

\vspace{2pt}
\hrule
\vspace{2pt}

\hfill
\begin{minipage}{0.92\linewidth}
\begin{algorithmic}[1]
\State \textbf{Input:} lexicon $\mathcal{L}$, parameters $\theta$
\State \textbf{Output:} token sequence $S$

\State \textbf{Step 1: Base Structure via \textsc{Merge}}
    \State Sample lexical items $V, D_1, D_2, N_1, N_2 \sim \mathcal{L}$
    \State $DP_{subj} = \textsc{Merge}(D_1, N_1)$, $DP_{obj} = \textsc{Merge}(D_2, N_2)$
    \State $V' = \textsc{Merge}(V, DP_{obj})$; $vP = \textsc{Merge}(DP_{subj}, V')$
    \State \textit{// Yields a hierarchical phrase structure with subject and object}

\State \textbf{Step 2: Functional Structure and \textsc{Agree}}
    \State Assign number feature $iNum \in \{sg, pl\}$ to $DP_{subj}$
    \State Create $T[uNum]$ and \textsc{Merge} with $vP$
    \State Set $uNum \leftarrow iNum$ via $\textsc{Agree}(T, DP_{subj})$
    \State Form $TP = [TP\ DP_{subj}\ T[uNum]\ vP]$
    \State \textit{// Yields a subject--verb agreement dependency encoded in the structure}

\State \textbf{Step 3: \textsc{Move} (Dependency Encoding)}
    \State Copy $DP_{subj}$ to clause-initial position
    \State Replace its original position with a trace: $t_{subj}$
    \State Form $TP = [TP\ DP_{subj}\ T\ [vP\ t_{subj}\ [V'\ V\ DP_{obj}]]]$

    \State Sample $wh \in \{+,-\}$
    \State \textsc{Merge} $C[wh]$ with $TP$
    \State \textbf{if} $wh = +$ \textbf{then}
        \State Select a $DP$ in $TP$ and mark it as $DP_{wh}$
        \State Copy $DP_{wh}$ to clause-initial position
        \State Replace its original position with a trace $t_{wh}$
        \State Form $CP = [CP\ DP_{wh}\ C\ [TP\ ...\ t_{wh}\ ...]]$
    \State \textbf{end if}
    \State \textit{// Yields a long-distance dependency between a moved element and its trace}

\State \textbf{Step 4: Linearization}
    \State Traverse the tree in pre-order
    \State Output brackets, nonterminal labels, features, and traces
    \State Remove lexical terminals ($V, N, D$) $\rightarrow S$
    \State \textit{// Yields a token sequence encoding structural relations without lexical content}

\end{algorithmic}
\end{minipage}

\vspace{2pt}
\hrule
\end{algorithm}

Pre-pretraining (PPT) on synthetic structures has emerged as a new learning paradigm for language models. 
Existing studies have reported that training models via next-token prediction on data possessing hierarchical structures—such as MIDI music, programming languages, or specific formal languages—can impart useful inductive biases, thereby improving the efficiency of subsequent natural language learning~\cite{papadimitriou-jurafsky-2020-learning,ri-tsuruoka-2022-pretraining,papadimitriou-jurafsky-2023-injecting}.

More recently, \citet{hu-etal-2025-circuits} advanced this paradigm by introducing the \emph{Expressivity Hypothesis}, which posits that a formal language conferring a helpful inductive bias should be hierarchically structured---either context-free or context-sensitive---and definable in C-RASP~\cite{yang2024counting}.
However, while \citeauthor{hu-etal-2025-circuits}'s approach successfully exercises the model's generic computational capacity, it primarily focuses on abstract structural expressivity rather than properties characteristic of natural language.
For instance, $k$-Shuffle Dyck defines dependencies purely through bracket-matching rules, allowing flexible crossing patterns but lacking the asymmetric, head-driven organization typically observed in natural language.
This raises the question of whether incorporating more natural-language-like structural properties into synthetic sequences could lead to more effective inductive biases.

In this work, we take the LAD/UG perspective as motivation: rather than deriving a formal grammar, we operationalize the structural properties that these operations are assumed to encode---hierarchical composition, feature-based dependencies, and long-distance displacement---as explicit sequence-level patterns for use in a PPT setting.

\section{Methods}
\label{sec:lad_for_llm}

The goal of LAD-inspired PPT is to inject an inductive bias that proactively restricts the model's hypothesis space to natural-language-like structure before standard pretraining begins.
To this end, we propose \textsc{MP-Struct}, a data generator designed to produce \emph{serialized structural representations}---token sequences that make explicit the hierarchical and dependency structure assumed to underlie natural language.
The generator is not intended to model natural language itself, nor does it derive sequences from a formal grammar.
Instead, it operationalizes three structural properties---hierarchical composition (\textsc{Merge}), feature-based agreement (\textsc{Agree}), and long-distance displacement (\textsc{Move})---as abstract sequence-level patterns, stripped of all lexical content.
 
\textsc{MP-Struct} generates data according to Algorithm~\ref{alg:mg_gen} in the following steps.

\paragraph{Step 1: Base Structure via \textsc{Merge}}
We sample lexical items from a lexicon and construct a $vP$ bottom-up using the \textsc{Merge} operation.
This procedure yields a recursive hierarchical structure—rather than a flat sequence—which forms the structural backbone of the generated data.
 
\paragraph{Step 2: Functional Structure and \textsc{Agree}}
We introduce a functional head $T$ with an uninterpretable number feature ($uNum$), and assign an interpretable number feature ($iNum$) to the subject $DP$.
The value of $uNum$ is then determined via agreement with the subject.
This step encodes feature-based dependencies within the hierarchical structure.
 
\paragraph{Step 3: \textsc{Move} (Dependency Encoding)}
We encode long-distance dependencies by copying elements to higher structural positions and replacing their original occurrences with traces.
Specifically, the subject $DP$ is copied to a higher position, forming a dependency between the copied element and its trace.
We optionally introduce a complementizer $C$ with a binary $wh$ feature; when $wh=+$, a $DP$ is selected, copied to a higher position, and linked to its original position via a trace.
These operations result in structured dependencies spanning multiple hierarchical levels.
 
\paragraph{Step 4: Linearization}
We traverse the derived tree and output a sequence consisting of structural brackets, nonterminal labels, features, and traces, while omitting lexical items.
This design isolates structural information from lexical content: by removing lexical tokens, the model cannot rely on surface co-occurrence patterns and is instead encouraged to process hierarchical structure and dependency relations directly.
In the context of pre-pretraining, this is intended to encourage the model to acquire linguistically motivated inductive biases independently of lexical semantics, which may transfer to subsequent natural language pretraining.

\section{Experiments}
\label{sec:experiments}
We test whether introducing linguistically motivated inductive biases through PPT can improve the efficiency of subsequent natural language learning.

\subsection{Experimental Setup}
\paragraph{Model and training protocol.}
We follow the blockwise learning paradigm of~\citet{hu-etal-2025-circuits} and use \texttt{Pythia-1B}~\cite{biderman2023pythia} as the base model.
Each run consists of (i) an optional PPT phase and (ii) a natural-language pretraining phase.
For pretraining (PT), we train on \texttt{C4}~\cite{Raffel2019ExploringTL} for $25{,}000$ optimization steps.
For all PPT conditions, we fix the budget for the synthetic PPT phase to $500$ steps, and subsequently we transfer the parameters to initialize the standard PT.
All experiments are conducted with three random seeds, and we report the mean over seeds.
The details of the training hyperparameters are provided in Appendix~\ref{appendix:detailed_settings}.

\paragraph{PPT corpora.}
For PPT, we pretrain on either (i) unstructured synthetic sequences (\textbf{Random}), (ii) formal languages with explicit recursion and/or crossing dependencies (\textbf{1-Dyck}, \textbf{$k$-Shuffle Dyck}), or (iii) our LAD-inspired structural representations (\textbf{\textsc{MP-Struct}}).
All synthetic sequences are tokenized with the \texttt{Pythia-1B} tokenizer.
Detailed generation hyperparameters are provided in Appendix~\ref{appendix:mp_strct_hyperparametrs}.

\subsection{Baselines}
We compare \textsc{MP-Struct} with the following baselines to isolate specific sources of efficiency gains:
\begin{itemize}[nosep]
  \item \textbf{Non-PPT:} Standard pretraining from random initialization. This serves as the baseline to quantify the absolute benefit of introducing any PPT phase.
  \item \textbf{Random:} An unstructured PPT control trained on i.i.d. uniformly sampled tokens. This verifies that gains are not merely due to additional gradient updates or data exposure, but specifically stem from \textit{structural} inductive bias.
  \item \textbf{1-Dyck:} A minimal recursion baseline representing pure context-free structure (definable in C-RASP). This tests the sufficiency of pure recursive nesting without the complexity of crossing dependencies.
  \item \textbf{$k$-Shuffle Dyck:} The current state-of-the-art formal bias~\cite{hu-etal-2025-circuits}. We specifically adopt the configuration with $k=64$, following the base configuration of \citet{hu-etal-2025-circuits}.
  It is a context-sensitive language yet remains definable in C-RASP, theoretically necessitating both Stack- and Queue-like memory operations.
\end{itemize}

\begin{figure}[t!]
 \centering
  \includegraphics[width=0.9\linewidth]{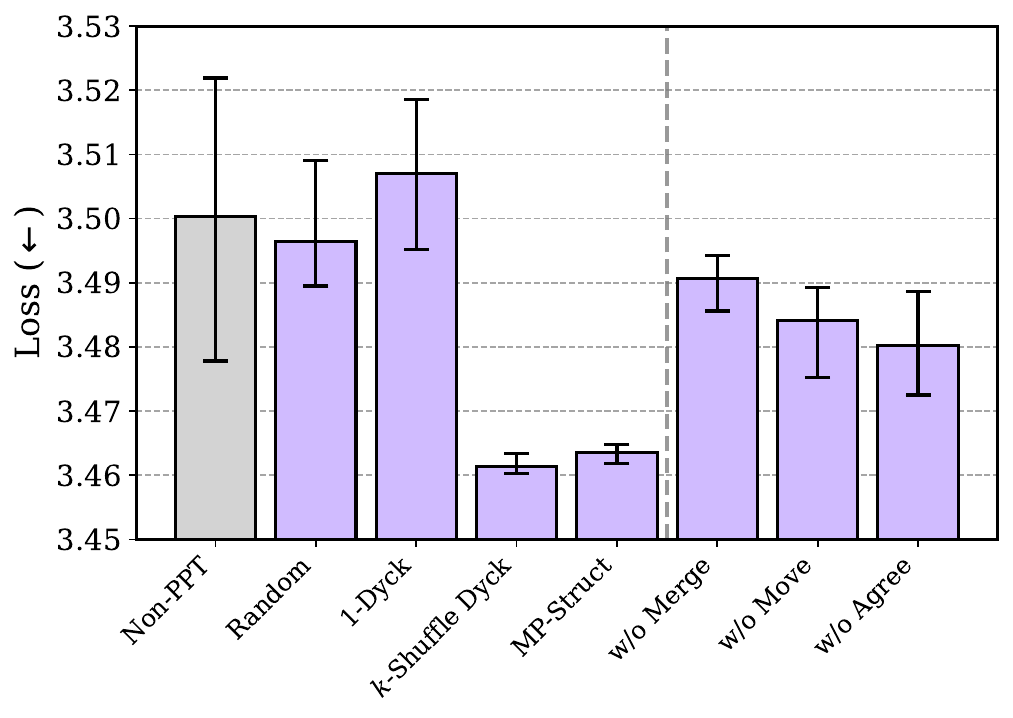}
 \caption{Comparison of C4 validation loss at 25,000 steps across different pre-pretraining conditions. The left section compares \textsc{MP-Struct} against baselines, while the right section (separated by the dashed line) presents an ablation study removing specific linguistic components (\textsc{Merge}, \textsc{Move}, \textsc{Agree}).}
 \label{fig:lm_loss}
\end{figure}

\begin{table}[t] 
    \centering
    \small 
    \setlength{\tabcolsep}{4pt} 
    \begin{tabular}{l c c c}
        \toprule
        \textbf{Model} & \textbf{BLiMP} & \textbf{MRS} & \textbf{Efficiency} \\
        \midrule
        Non-PPT & 0.758 & -- & -- \\
        Random & 0.761 & -- & -- \\
        1-Dyck & 0.759 & -- & -- \\
        $k$-Shuffle Dyck & \textbf{0.764}$^\dagger$ & 15.6 $\pm 3.31$ & 0.29 $\pm 0.068$ \\
        \textsc{MP-Struct} & 0.755 & 15.3 $\pm 2.86$ & 0.29 $\pm 0.057$  \\ \midrule
        \textsc{MP-Struct} Core & \textbf{0.764}$^\dagger$ & \textbf{16.2} $\pm 2.97$ & \textbf{0.31} $\pm 0.055$ \\
        \bottomrule
    \end{tabular}
    \caption{
        BLiMP, MRS, and Efficiency Gain (Efficiency).
        $^\dagger$ indicates a significant difference from Non-PPT at the 5\% level.
        ``--'' indicates that the condition did not improve upon Non-PPT (i.e., yielded negative MRS and Efficiency values), and is therefore excluded from the efficiency comparison.
    }
    \label{tab:main_results}
\end{table}

\subsection{Evaluation Metrics}
Following prior work on PPT~\cite{hu-etal-2025-circuits}, we evaluate whether improvements in learning efficiency are achieved without degrading the quality of the acquired grammar.
\begin{itemize}
    \item \textbf{Learning Efficiency:} We use two metrics from \citet{hu-etal-2025-circuits}. Let $y_1$ be the number of PT steps for the Non-PPT baseline, $x$ the number of PPT steps, and $y_2$ the number of PT steps at which the PPT model first matches the loss of Non-PPT at $y_1$.
    \textbf{Marginal Rate of Substitution (MRS)} measures how many PT steps are saved per PPT step:
    \begin{equation}
        \text{MRS} = \frac{y_1 - y_2}{x}
    \end{equation}
    \textbf{Efficiency Gain} measures the reduction in total training steps required to reach matched performance:
    \begin{equation}
        \text{Efficiency Gain} = 1 - \frac{y_2 + x}{y_1}
    \end{equation}
    A concrete calculation example is provided in Appendix~\ref{appendix:efficiency_metrics}.
    \item \textbf{Grammatical Generalization:} We use BLiMP~\cite{warstadt-etal-2020-blimp-benchmark}, which evaluates English grammar using minimal pairs.
\end{itemize}

\subsection{Results}
\label{sec:results}
Figure~\ref{fig:lm_loss} presents the C4 training loss after 25,000 steps, and Table~\ref{tab:main_results} summarizes the grammatical generalization and the learning efficiency.

\paragraph{1. Learning Efficiency Gains from \textsc{MP-Struct}.}
\textsc{MP-Struct} outperforms both the Non-PPT baseline and the unstructured Random control in terms of best loss (Figure~\ref{fig:lm_loss}, left).
Quantitatively, \textsc{MP-Struct} achieves an MRS of 15.3 on average relative to Non-PPT, corresponding to an average efficiency gain of 29\% (up to 35\%).
In contrast, the 1-Dyck baseline does not yield consistent improvements, suggesting that recursive structure alone may not be sufficient to improve learning efficiency.

\paragraph{2. Synergy of Linguistic Operations.}
To isolate the drivers of this efficiency gain, we conducted an ablation study (Figure~\ref{fig:lm_loss}, right).
The results reveal that removing any single component of the generator---\textsc{Merge}, \textsc{Agree}, or \textsc{Move}---results in a worse final loss compared to the full \textsc{MP-Struct} model.
This confirms that the efficiency gain is not driven by any isolated feature but by the \textit{synergistic interaction} of hierarchical phrase structure (\textsc{Merge}) and functional dependencies (\textsc{Agree}/\textsc{Move}), validating the theoretical design of Algorithm~\ref{alg:mg_gen}.

\paragraph{3. Comparable Performance to $k$-Shuffle Dyck.}
\textsc{MP-Struct} achieves an average efficiency gain of 29\%, comparable to the strong baseline $k$-Shuffle Dyck (29\%).
While $k$-Shuffle Dyck shows marginally higher BLiMP scores, \textsc{MP-Struct} yields a BLiMP score comparable to the Non-PPT baseline ($0.755$ vs.\ $0.758$).
This suggests that the induced inductive bias primarily facilitates learning efficiency rather than improving final grammatical generalization.
These results indicate that linguistically motivated inductive biases can serve as an alternative to high-expressivity baselines for improving token efficiency, and motivate further analysis of the factors underlying these gains (\S\ref{sec:analysis_mechanism}).

\section{Analysis I: Quality of Inductive Bias}
\label{sec:analysis_bias}

\begin{figure}[t]
 \centering
  \includegraphics[width=0.65\linewidth]{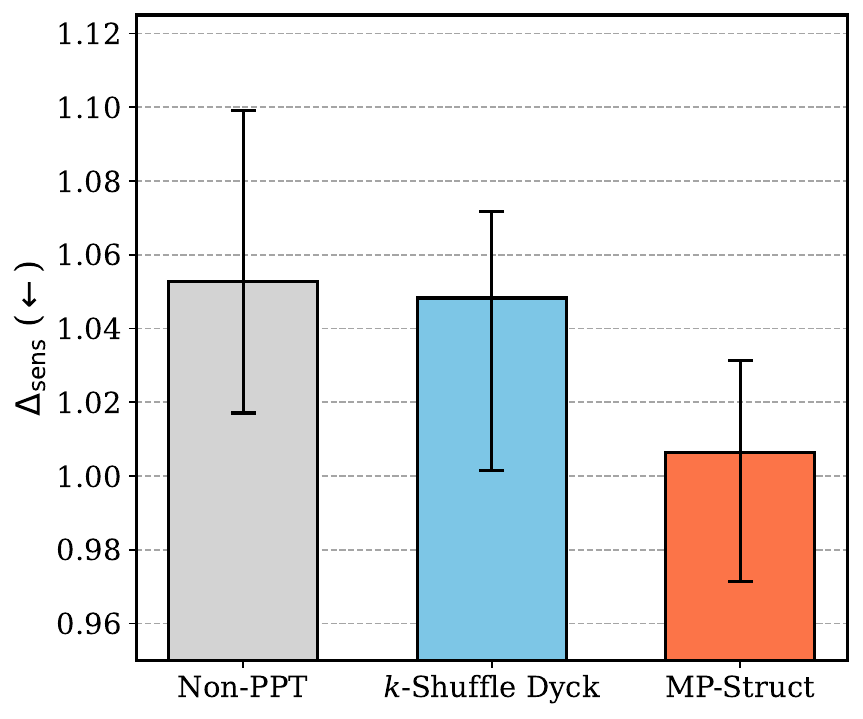}
    \caption{
        \textbf{Robustness against semantic perturbation} ($\Delta_{\text{sens}} = \mathcal{L}_{\text{JW}} - \mathcal{L}_{\text{NL}}$).
        This metric quantifies the performance gap between semantic-free Jabberwocky inputs and natural language, where lower values indicate less reliance on lexical co-occurrence.
    }
    \label{fig:jw_nl_delta}
\end{figure}

\begin{figure}[t]
    \centering
    \begin{subfigure}[b]{0.32\textwidth}
        \centering
        \includegraphics[width=\textwidth]{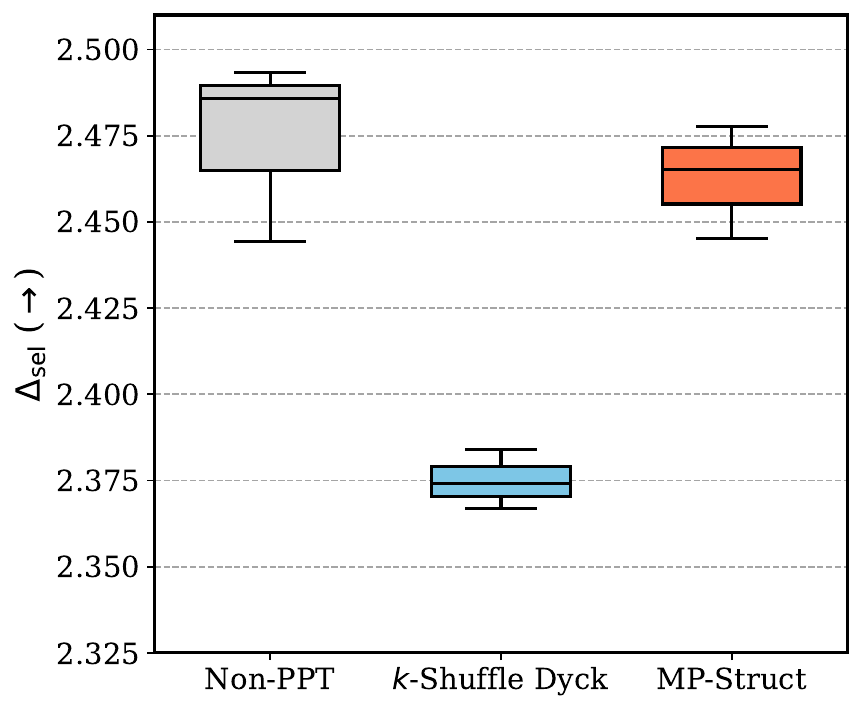}
        \caption{\textsc{Shuffle}}
        \label{fig:shuffle}
    \end{subfigure}
    \hfill 
    \begin{subfigure}[b]{0.32\textwidth}
        \centering
        \includegraphics[width=\textwidth]{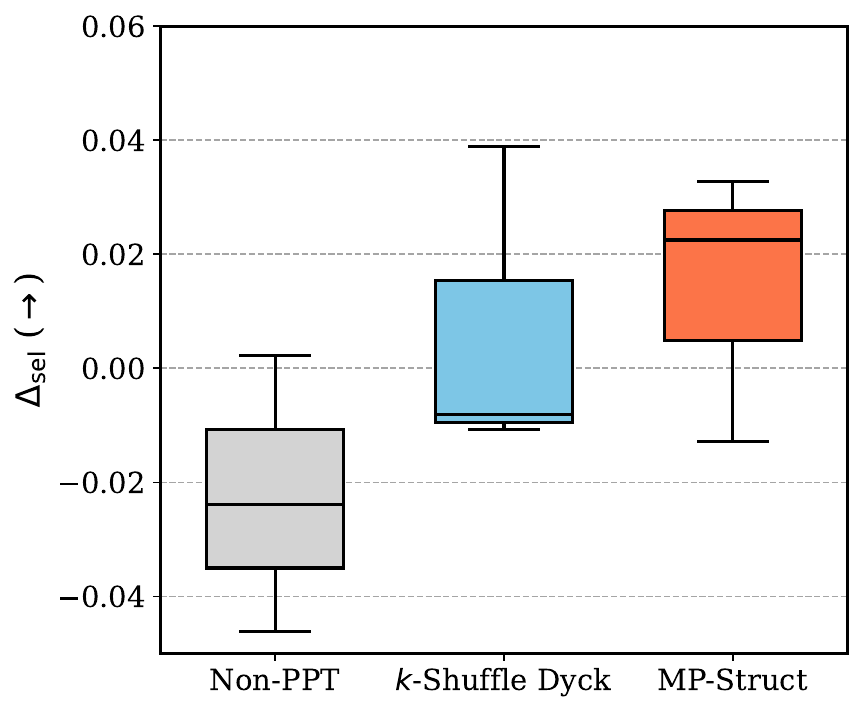}
        \caption{\textsc{Reverse}}
        \label{fig:reverse}
    \end{subfigure}
    \hfill
    \begin{subfigure}[b]{0.32\textwidth}
        \centering
        \includegraphics[width=\textwidth]{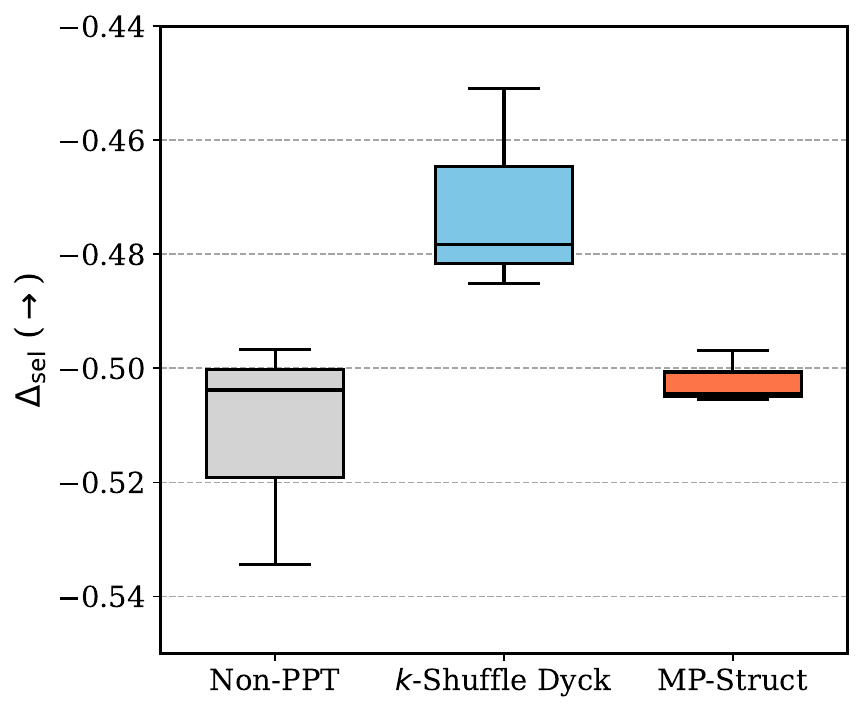}
        \caption{\textsc{Hop}}
        \label{fig:hop}
    \end{subfigure}

    \caption{
        \textbf{Structural selectivity} ($\Delta_{\text{sel}} = \mathcal{L}_{\text{Imp}} - \mathcal{L}_{\text{NL}}$) across three impossible language conditions.
        Positive values indicate a human-like preference for natural linguistic constraints over impossible distortions.
        }  
    \label{fig:impossible_delta_3plots}        
\end{figure}

Having established the efficiency of \textsc{MP-Struct}, we now investigate the \emph{nature} of the acquired biases.

\subsection{Validation of Structural Robustness}
\label{subsec:skelton}

To examine whether the observed efficiency gains are associated with improved structural processing, we perform a \emph{Jabberwocky (JW)} meaning attenuation analysis~\cite{Carroll1871}.
This analysis is intended to probe the extent to which models rely on structural information independently of lexical semantics.
We construct JW variants of both the training (C4) and evaluation (Wikitext~\cite{merity2016pointer}) corpora by randomly replacing content words while preserving function words, punctuation, and word order, conditioned on fine-grained POS tags to preserve morphology.\footnote{We provide the details of data construction in Appendix~\ref{appendix:jabberwocky_construction}}

The engineering significance of this analysis lies in quantifying the \emph{disentanglement} of syntactic processing from semantic correlation.
Current LLMs often rely on lexical co-occurrence statistics to minimize loss~\cite{dziri-etal-2023-faith,berglund-etal-2023-reversal}.
However, a robust language learner is expected to maintain predictive performance even when semantic cues are reduced, relying instead on structural regularities~\cite{gulordava-etal-2018-colorless}.

To assess this, we train separate models under two conditions: natural language (NL) and its Jabberwocky (JW) counterpart, and evaluate each model on the corresponding data (i.e., NL$\rightarrow$NL and JW$\rightarrow$JW).
We then compare their losses to quantify sensitivity to semantic information.

We define sensitivity as $\Delta_{\text{sens}} = \mathcal{L}_{\text{JW}} - \mathcal{L}_{\text{NL}}$, where $\mathcal{L}_{\text{NL}}$ and $\mathcal{L}_{\text{JW}}$ denote the losses obtained under the NL$\rightarrow$NL and JW$\rightarrow$JW conditions, respectively.
A smaller $\Delta_{\text{sens}}$ indicates that performance degrades less when semantic information is attenuated, which may suggest greater reliance on structural information.

\paragraph{Results.}
Figure~\ref{fig:jw_nl_delta} shows that while Non-PPT relies heavily on semantic co-occurrence, \textsc{MP-Struct} achieves a lower $\Delta_{\text{sens}}$ than the strong baseline $k$-Shuffle Dyck.

While $k$-Shuffle Dyck theoretically requires memory operations capable of tracking long-distance dependencies, its symbol types lack explicit cues that differentiate their structural roles, potentially leaving multiple plausible antecedents and leading to higher dependency retrieval ambiguity.
In contrast, \textsc{MP-Struct} introduces distinct structural markers (e.g., functional categories such as $T$ and $C$), which may help reduce ambiguity in identifying relevant dependencies.

One possible interpretation is that such explicit cues make it easier for the model to rely on structural information when semantic content is attenuated.
Accordingly, the improved robustness of \textsc{MP-Struct} under meaning attenuation may reflect a greater reliance on structural regularities, rather than lexical co-occurrence alone.
To further investigate this hypothesis, we analyze the factors underlying these efficiency gains in \S\ref{sec:analysis_mechanism}.

\subsection{Resistance to Impossible Languages}
\label{sec:impossible_analysis}

We next examine whether the initialization induced by \textsc{MP-Struct} tends to align with constraints characteristic of human language (UG) by probing learning behavior on \emph{impossible languages}.
Following the protocol of \citet{kallini-etal-2024-mission}, we construct synthetically perturbed corpora by applying deterministic transformations to the NL training data. 
Specifically, we adopt the following implementations from their framework to violate specific linguistic universals while retaining statistical regularities:\footnote{We provide the details of data construction in Appendix~\ref{appendix:impossible_languages}.}

\begin{itemize}[nosep]
    \item \textbf{\textsc{Shuffle}}: We use \textsc{DeterministicShuffle} with a window size of $s=21$. This operation permutes tokens deterministically within a fixed local window, destroying local $n$-gram statistics and syntactic constituency while preserving the global bag-of-words distribution.
    \item \textbf{\textsc{Reverse}}: We use \textsc{FullReverse}, which reverses the token order of the entire sequence. While computationally deterministic (requiring a stack), this transformation violates the incremental, left-to-right processing constraint fundamental to human language performance.
    \item \textbf{\textsc{Hop}}: We use \textsc{WordHop} (specifically with a 4-word shift). This transformation introduces a dependency based on linear counting: a functional marker is placed at a fixed linear distance (four words) after its associated verb. This mimics ``impossible'' grammatical rules that rely on counting word positions rather than structural configurations.
\end{itemize}

We quantify the model's preference for natural constraints using \emph{structural selectivity}: $\Delta_{\text{sel}} = L_{\text{Imp}} - L_{\text{NL}}$. 
\emph{A larger (positive) $\Delta_{\text{sel}}$} indicates that the model finds natural language significantly easier to learn than impossible languages, implying a bias toward human-like structures.

\paragraph{Results and Discussion.}

Figure~\ref{fig:impossible_delta_3plots} presents the structural selectivity scores ($\Delta_{\text{sel}}$) at 25,000 pretraining steps across seeds.
In the \textsc{Shuffle} condition (Fig.~\ref{fig:impossible_delta_3plots}a), all models exhibit consistently positive $\Delta_{\text{sel}}$, indicating that a preference for local structural coherence is broadly shared regardless of PPT condition.
In the \textsc{Hop} condition (Fig.~\ref{fig:impossible_delta_3plots}c), $\Delta_{\text{sel}}$ values are negative across conditions, reflecting the Transformer's inherent capacity to capture long-distance dependencies.
These two conditions do not clearly differentiate the PPT conditions from each other or from Non-PPT.
 
The most informative divergence appears in the \textsc{Reverse} condition (Fig.~\ref{fig:impossible_delta_3plots}b).
Here, $k$-Shuffle Dyck yields $\Delta_{\text{sel}} \approx 0$, suggesting that it incentivizes the acquisition of generic processing strategies capable of handling sequences in any direction---an excessive flexibility that lacks linguistic structural constraints.
In contrast, \textsc{MP-Struct} maintains a clearly positive $\Delta_{\text{sel}}$, indicating resistance to reversed sequences.
This resistance is not incidental: the sequences generated by \textsc{MP-Struct} encode a fundamentally directional structure, in which displacement consistently targets structurally higher positions to the left, imposing a directional asymmetry on the linearized sequence.
Reversing the input string directly violates these directional biases, making reversed sequences genuinely harder to process for a model that has internalized them.
We interpret this as evidence that \textsc{MP-Struct} instills a directionally asymmetric inductive bias, consistent with the LAD-inspired design goal of favoring natural-language-like structure.

\section{Analysis II: Drivers of Efficiency}
\label{sec:analysis_mechanism}

 \begin{table*}[t]
    \centering
    \small
    \begin{tabular}{l p{10.0cm}}
    \toprule
    \textbf{Condition} & \textbf{Example Sequence} \\
    \midrule
    
    \textbf{1. Generic $k$-SD} & 
    \texttt{\skel{[0} \depA{(1} \depA{(2} \depA{(4} \skel{]0} \depA{)4} \depA{)1} \skel{[0} \depA{)2} \skel{]0}} \\
    \textit{(Random Mix)} &
     \\
    \midrule
    
    \textbf{2. \textsc{MP-Struct Core}} & 
    \texttt{\skel{[0} \head{H\_C} \depA{(1} \skel{[0} \head{H\_T} \depA{(2} \skel{[0} \depA{(4} \skel{]0} \depA{)4} \skel{]0} \depA{)2} \skel{]0} \depA{)1} \skel{[0} \head{H\_V} \skel{]0} \skel{]0}} \\
    \textit{(Constrained Mix)} & 
     \\
    \bottomrule
    \end{tabular}
    \caption{
        \textbf{Decomposition of Abstract Generative Conditions.}
        We contrast \textbf{Generic $k$-SD} with \textbf{\textsc{MP-Struct Core}}, which introduces diverse functional heads as explicit landmarks.
        Both conditions share the same set of dependency types; they differ in whether dependencies are randomly interleaved (Generic $k$-SD) or organized within a fixed hierarchical topology with landmark tokens adjacent to each dependency site (\textsc{MP-Struct Core}).
        Colors: \skel{Hierarchy [ ]}, \depA{Dependency ( )}, and \head{Heads}.
    }
    \label{tab:search_complexity_analysis}
\end{table*}

\begin{algorithm}[t]
\caption{Data Generation Procedure (\textsc{MP-Struct Core})}
\label{alg:enhanced_constrained_gen}
\centering
\footnotesize
\hrule \vspace{2pt}
\begin{minipage}{0.98\linewidth}
\textbf{Notation:} $[\,\cdot\,]$/$(\,\cdot\,)$: structural/dependency bracket pair, $T/C$: functional heads, $t$: trace, $\mathrm{AGR} \in \{\mathrm{AGR}_A, \mathrm{AGR}_B\}$: agreement dependency (subject--verb number agreement), $\mathrm{SEL}$: selectional dependency (head selects its complement, e.g., $D$ selects $N$), $\mathrm{MOVE}$: movement dependency, $H_X$: head token for layer $X \in \{CP, TP, VP\}$, $wh \in \{+,-\}$.
\end{minipage}
\vspace{2pt}
\hrule
\vspace{2pt}
\hfill
\begin{minipage}{0.92\linewidth}
\begin{algorithmic}[1]
\State \textbf{Input:} sequence length $L$
\State \textbf{Output:} token sequence $S$
\State \textbf{Step 1: Base Structure via \textsc{Merge}}
    \State Sample $wh \in \{+,-\}$ and $\mathrm{AGR} \in \{\mathrm{AGR}_A, \mathrm{AGR}_B\}$
    \State Construct $vP$ containing head $H_{VP}$, a subject slot (trace $t$ if $wh{=}+$, else empty), and one object linked via $\mathrm{SEL}$ dependency
    \State Shuffle $vP$-internal elements to vary surface order
    \State \textit{// Yields an abstract verb phrase with a selectional dependency between head and object}

\State \textbf{Step 2: Functional Structure and \textsc{Agree}}
    \State Assign $\mathrm{AGR}$ feature to $DP_{subj}$; create $T[H_{TP},\,\mathrm{AGR}]$ and \textsc{Merge} with $vP$
    \State \textbf{if} $wh = +$ \textbf{then} leave Spec-TP empty \textbf{else} place $DP_{subj}$ at Spec-TP \textbf{end if}
    \State Form $TP = [TP\ \mathrm{Spec\text{-}TP}\ T\ vP]$
    \State \textit{// Yields a subject--verb agreement dependency marked by landmark $H_{TP}$}

\State \textbf{Step 3: \textsc{Move} (Dependency Encoding)}
    \State \textbf{if} $wh = +$ \textbf{then}
        \State Copy $DP_{subj}$ to Spec-CP; leave trace $t_{subj}$ at its $vP$-internal position
        \State Attach licensor $\mathrm{MOVE}$ to $C$; form $CP = [CP\ DP_{subj}\ C[H_{CP},\,\mathrm{MOVE}]\ TP]$
    \State \textbf{else}
        \State Form $CP = [CP\ \text{(empty)}\ C[H_{CP}]\ TP]$
    \State \textbf{end if}
    \State \textit{// Yields a long-distance movement dependency anchored by landmark $H_{CP}$}

\State \textbf{Step 4: Linearization}
    \State Traverse the tree in pre-order
    \State Output brackets, dependency markers, and traces $\rightarrow S$
    \State Repeat Steps 1--4 and concatenate until $|S| \geq L$
    \State \textit{// Yields a token sequence where each dependency site is marked by an unambiguous landmark}
\end{algorithmic}
\end{minipage}
\vspace{2pt}
\hrule
\end{algorithm}
\begin{figure}[t!]
 \centering
  \includegraphics[width=0.9\linewidth]{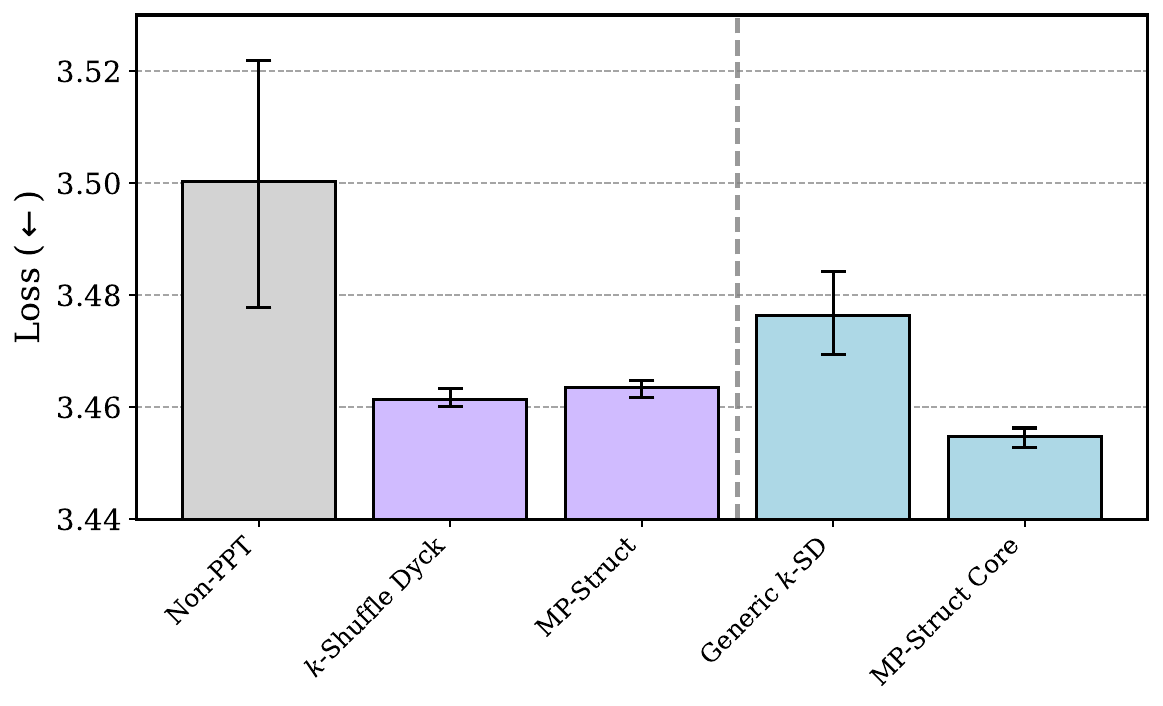}
 \caption{Comparison of C4 loss at 25,000 steps across abstract conditions.}
 \label{fig:lm_loss_abstract}
\end{figure}

The results in \S\ref{subsec:skelton} suggest that the observed efficiency gains may be related to differences in how structural information is represented, particularly the availability of explicit cues for dependency resolution.
This raises a more specific question: \textit{are these gains driven solely by structural expressivity (i.e., C-RASP definability), or by how dependencies are organized and made accessible to the model?}

\subsection{Decomposition and Abstract Conditions}
\label{subsec:decomposition}

To better isolate the factors contributing to learning efficiency, we construct a controlled experimental setting in which the set of structural operations is held constant across conditions.
Specifically, both conditions share the same primitive operations as \textsc{MP-Struct}: one type of recursive structure and four types of functional dependencies (see Appendix~\ref{appendix:complexity_calibration} for details).
The only factor varied is how these components are organized, allowing us to attribute any difference in efficiency directly to the organization of dependencies rather than to their number or type.
Within this controlled setting, we analyze how the \emph{organization} of dependencies affects learning, focusing on what we term \emph{dependency identification ambiguity}.
 
Dependency identification ambiguity refers to the extent to which a dependency endpoint (e.g., \texttt{\depA{)1}}) provides sufficient cues to uniquely identify its corresponding start point (e.g., \texttt{\depA{(1}}).
When such cues are limited, multiple candidate antecedents remain plausible, leading to high ambiguity.
Conversely, when structural markers are present near relevant positions, the set of candidates can be sharply constrained, resulting in lower ambiguity.

\begin{figure}[t!]
 \centering
  \includegraphics[width=1.0\linewidth]{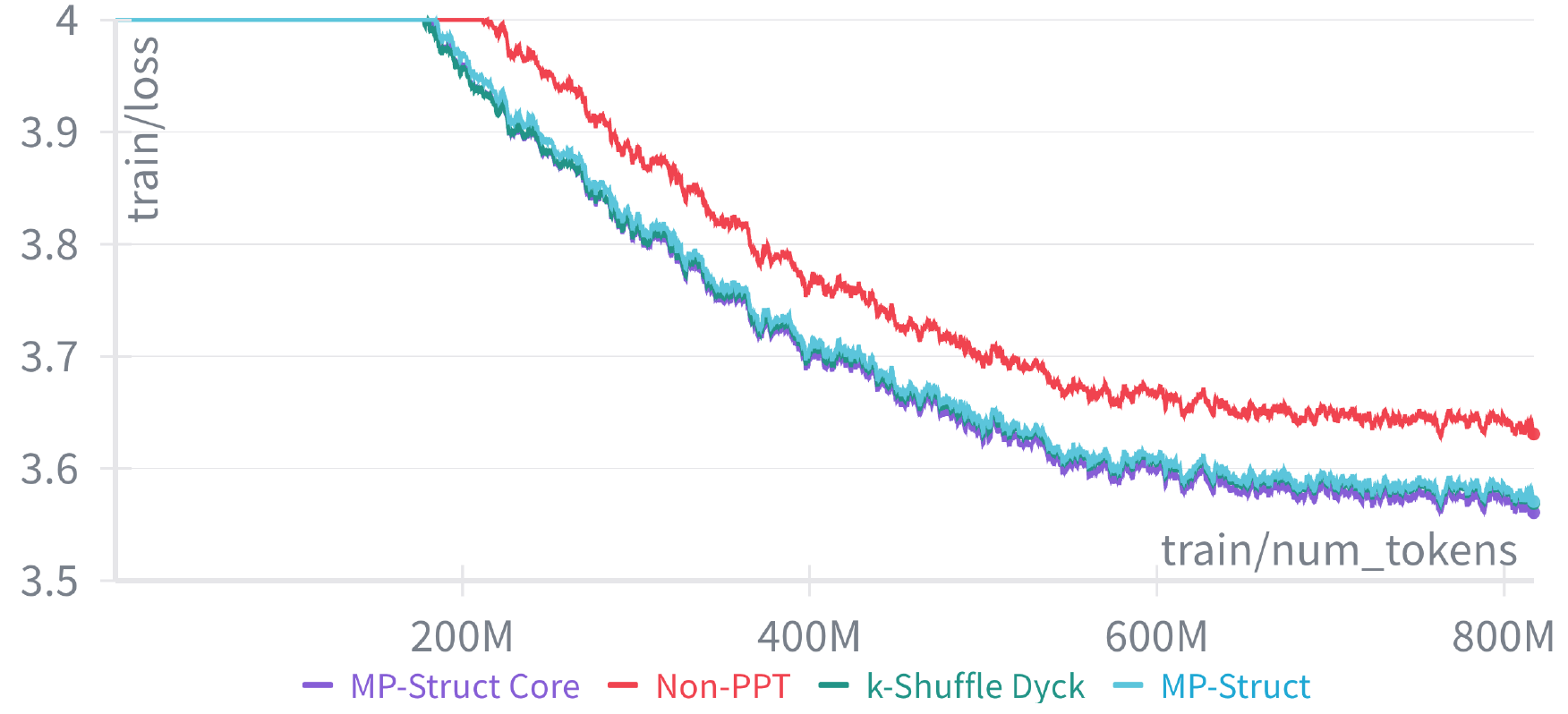}
 \caption{The trajectory of the training loss.}
 \label{fig:loss_curve}
\end{figure}

Based on this perspective, we define two contrasting conditions:
 \begin{itemize}[nosep]
    \item \textbf{1. Generic $k$-SD:}
    A condition where the same primitive components are randomly interleaved.
    As illustrated in Table~\ref{tab:search_complexity_analysis}, multiple dependency start points of the same type (e.g., two instances of \texttt{\depA{(1}}) may appear in an unstructured manner, providing weak cues for identifying which one corresponds to a given endpoint (e.g., \texttt{\depA{)1}}).
    As a result, multiple candidates remain plausible, leading to \textbf{high dependency identification ambiguity}.
 
    \item \textbf{2. \textsc{MP-Struct Core}:}
    A distillation of \textsc{MP-Struct} into a pure abstract formal language, designed to preserve its two key structural properties while eliminating lexical content (see Algorithm~\ref{alg:enhanced_constrained_gen} for the full generation procedure\footnote{We provide supplementary detail on the mapping from \textsc{MP-Struct} to its abstract counterpart in Appendix~\ref{appendix:core_generation}.}):
 
    \begin{itemize}[nosep]
        \item \textbf{Fixed Topology.}
        The derivation strictly follows the hierarchical path $CP \to TP \to vP$, enforcing the same recursive subordination as \textsc{MP-Struct}.
        Unlike Generic $k$-SD, where dependencies are randomly interleaved, every dependency in \textsc{MP-Struct Core} is generated within its designated structural domain.
 
        \item \textbf{Abstract Landmarks.}
        The functional heads of \textsc{MP-Struct} ($C$, $T$, $v$) are replaced by distinct abstract tokens $H_{CP}$, $H_{TP}$, $H_{VP}$, which are systematically placed adjacent to their associated dependency brackets.
        This ensures that each dependency site is marked by an unambiguous, consistent cue, mirroring the role of functional heads without introducing lexical noise.
    \end{itemize}
 
    Together, these properties result in \textbf{low dependency identification ambiguity}: the head tokens serve as explicit landmarks that sharply localize the relevant search space for each dependency.
\end{itemize}

\subsection{Analysis of Efficiency Factors}
\label{subsec:efficiency_factors}

Figure~\ref{fig:lm_loss_abstract} summarizes the results of our abstraction study.\footnote{The trajectory of the LM loss is provided in Figure~\ref{fig:loss_curve}.}
Among the abstract conditions, \textsc{MP-Struct Core} outperforms Generic $k$-SD in terms of token efficiency.
Quantitatively, \textsc{MP-Struct Core} achieves an MRS of 16.2 and an average Efficiency Gain of 31\% (up to 37\%) as shown in Table~\ref{tab:main_results}.
These values are higher than those of $k$-Shuffle Dyck (MRS: 15.6, Efficiency: 29\%).
These results suggest that differences in how structural components are organized may have a substantial impact on learning efficiency, even when the underlying set of operations---one type of recursive structure and four types of functional dependencies---is held constant across conditions.
 
In line with the design described in \S\ref{subsec:decomposition}, the two conditions differ primarily in the degree of dependency identification ambiguity.
Generic $k$-SD exhibits higher ambiguity due to the lack of explicit cues for identifying dependency relations, whereas \textsc{MP-Struct Core} provides more localized structural markers that help constrain the set of plausible antecedents.
From this perspective, reduced dependency identification ambiguity may contribute to more efficient learning, consistent with the hypothesis that structural accessibility—beyond expressivity alone—plays a role in effective PPT design.

\subsection{Theoretical Implication}
A key implication of this analysis concerns the relationship to the expressivity hypothesis \cite{hu-etal-2025-circuits}.
This hypothesis posits that a formal language conferring a helpful inductive bias should be hierarchically structured and definable in C-RASP~\cite{yang2024counting}, the latter serving as a formal lower bound on what future-masked soft attention transformers can express.
 
However, \textsc{MP-Struct Core} is not definable in C-RASP.
The generator enforces a strict adjacency constraint: functional head tokens (e.g., $H_C$) are systematically placed immediately before their associated dependency brackets, requiring a predicate that jointly references both the head position and the dependency position.
C-RASP, being a restriction of FO(M) that permits only single-index predicates per quantifier \cite{yang2024counting}, cannot express such a two-position constraint.
While \citet{hu-etal-2025-circuits} suggest that C-RASP-definability is a desirable property for effective PPT languages, this is framed as a tendency rather than a strict requirement---their results show only that C-RASP-definable languages \emph{generally} achieve equal or better performance, not that non-C-RASP-definable languages necessarily fail.
Consistent with this, \textsc{MP-Struct Core} achieves higher efficiency than $k$-Shuffle Dyck despite not being C-RASP-definable.
 
Taken together, these results suggest that C-RASP-definability is neither necessary nor sufficient for effective PPT, and that the organization of dependencies---specifically, the availability of explicit structural cues that reduce retrieval ambiguity---is a key complementary factor.

\section{Conclusion}
We proposed \emph{LAD-inspired PPT}, a pre-pretraining framework in which \textsc{MP-Struct}---a formal language taking cues from the Minimalist Program---injects linguistically motivated inductive biases before standard pretraining.
Our results show that such biases improve learning efficiency comparably to strong formal baselines, while additionally instilling a directionally asymmetric inductive bias, as evidenced by greater resistance to directionally reversed sequences compared to $k$-Shuffle Dyck.
Through controlled analyses, we find that the expressivity hypothesis alone does not fully account for these gains: \textsc{MP-Struct Core} outperforms $k$-Shuffle Dyck despite not being definable in C-RASP.
Instead, our analysis points to \emph{functional landmarks}---explicit structural cues that reduce dependency identification ambiguity---as a key complementary factor, suggesting that effective PPT design depends not only on formal expressivity but also on how structural information is organized to support efficient dependency retrieval.

\section*{Limitations}
While our results provide compelling evidence for the efficacy of linguistically motivated PPT, several limitations remain.

\paragraph{Scale and Architecture.}
Our experiments were conducted primarily on the Pythia-1B model~\cite{biderman2023pythia}. 
While we observed consistent trends across seed runs and smaller scales, it remains to be verified whether the efficiency gains of \textsc{MP-Struct} scale linearly to significantly larger models (e.g., 7B or 70B parameters) or alternative architectures (e.g., State-Space Models).

\paragraph{Monolingual Evaluation.}
We evaluated grammatical generalization using BLiMP~\cite{warstadt-etal-2020-blimp-benchmark}, which is limited to English. Although \textsc{MP-Struct} is designed based on Universal Grammar principles (e.g., \textsc{Merge} and \textsc{Move}) assumed to be language-universal, our current validation does not explicitly confirm improved acquisition efficiency for typologically distinct languages (e.g., head-final languages like Japanese or morphologically rich languages).

\paragraph{Operationalization of Dependency Identification Ambiguity.}
While we use dependency identification ambiguity as an explanatory construct, it currently lacks a formal, corpus-independent definition that would allow quantitative comparison across arbitrary languages.
For instance, it remains unclear whether $k$-Shuffle Dyck, C4, or other corpora exhibit higher or lower ambiguity than the conditions studied in \S\ref{sec:analysis_mechanism}, limiting the generalizability of our claims.
Furthermore, the comparison between Generic $k$-SD and \textsc{MP-Struct Core} may not isolate ambiguity as cleanly as intended: introducing landmark tokens not only reduces retrieval ambiguity but also increases vocabulary size, which may alter other properties of the language such as entropy.
Disentangling these confounds---for example, by controlling for unigram entropy or vocabulary size---remains an important direction for future work.

\section*{Acknowledgments}
We thank the anonymous reviewers for their helpful comments and suggestions.
This work was supported by JSPS KAKENHI Grant Number JP24H00087, Grant-in-Aid for JSPS Fellows JP24KJ0800, JST BOOST Grant Number JPMJBY24B2, JST CREST Grant Number JPMJCR2565, and JST PRESTO Grant Number JPMJPR21C2.

\bibliography{custom}

\clearpage
\appendix

\section{Training Configurations}
\label{appendix:detailed_settings}
Table~\ref{tab:train_hparams} shows the shared training hyperparameters for PPT and PT, following~\cite{hu-etal-2025-circuits}.
For the experiment, a single NVIDIA RTX 6000 Ada (48GB) GPU was used, and the training time for each run was approximately 20 hours.

\begin{table}[h!]
\centering
\small
\begin{tabular}{l c}
\toprule
\textbf{Hyperparameter} & \textbf{Value} \\
\midrule
Batch size & 16 \\
Gradient accumulation & 2 \\
Effective batch size & 32 \\
Max sequence length & 1024 \\
Learning rate & $5\times10^{-4}$ \\
LR schedule & Cosine with warmup \\
Min.\ LR & $5\times10^{-5}$ \\
Warmup steps & 1000 \\
Weight decay & 0.1 \\
Gradient clipping & 1.0 \\
Optimizer & AdamW \\
$\beta_1,\beta_2$ & 0.9, 0.999 \\
$\epsilon$ & $10^{-6}$ \\
Mixed precision & bf16 \\
\bottomrule
\end{tabular}
\caption{Shared training hyperparameters for PPT and PT (identical across all models), following~\cite{hu-etal-2025-circuits} except for max sequence length.}
\label{tab:train_hparams}
\end{table}

\section{Hyperparameters for \textsc{MP-Struct}}
\label{appendix:mp_strct_hyperparametrs}
Table~\ref{tab:mpstruct_hparams} shows the hyperparameters for \textsc{MP-Struct}.

\begin{table}[h!]
\centering
\small
\begin{tabular}{l c}
\toprule
\textbf{Hyperparameter} & \textbf{Value} \\
\midrule
\# sequences ($n$) & 100{,}000 \\
Max length & 1024 \\
Strip lexical items & True \\
EPP on $T$ & True \\
$P(C[+wh])$ & 0.2 \\
$P(DP[-wh])$ & 0.2 \\
Number prior ($P(\mathrm{sg})$) & 0.5 \\
Agreement-match ratio & 1.0 \\
\bottomrule
\end{tabular}
\caption{\textsc{MP-Struct} corpus generation hyperparameters.}
\label{tab:mpstruct_hparams}
\end{table}

\section{Examples used in pre-pretraining}
\label{appendix:ppt_corpora}
Table~\ref{tab:example_ppt} shows the examples used in pre-pretraining.

\begin{table}[h!]
\centering
\begin{tabularx}{\columnwidth}{X c}
\toprule
\textbf{Language} & \textbf{Example} \\
\midrule
\small{1-Dyck} & \texttt{(\,(\,(\,)\,)\,)} \\
\small{$k$-Shuffle Dyck} & \texttt{(\,\textcolor{blue}{[}\,\textcolor{red}{\{}\,\textcolor{blue}{]}\,)\,\textcolor{red}{\}}} \\
\small{\textsc{MP-Struct}} &
\parbox[t]{0.65\columnwidth}{%
    \ttfamily\small
[ CP [ C ] [ [ TP [ [ DP[Num:pl] [ D ] [ N ] ] ] [ T(+EPP,uNum:pl) ] [ [ VP V [ [ DP[Num:pl] [ D ] [ N ] ] ] [ TR[DP] ] ] ] ] ] ]
} \\
\textsc{w/o Merge} &
\parbox[t]{0.65\columnwidth}{%
    \ttfamily\small
C [ D N DP[Num:pl] [-wh] ] T(+EPP,uNum:pl) V D N DP[Num:pl] TR[-wh]
} \\
\textsc{w/o Move} &
\parbox[t]{0.65\columnwidth}{%
    \ttfamily\small
[ CP [ C ] [ [ TP [ T(+EPP,uNum:pl) ] [ [ VP V [ [ DP[Num:pl] [ D ] [ N ] ] [-wh] ] [ [ DP[Num:pl] [ D ] [ N ] ] ] ] ] ] ] ]
} \\
\textsc{w/o Agree} &
\parbox[t]{0.65\columnwidth}{%
    \ttfamily\small
[ CP [ C ] [ [ TP [ [ DP[Num:pl] [ D ] [ N ] ] [-wh] ] [ T(+EPP,uNum:u) ] [ [ VP V [ [ DP[Num:pl] [ D ] [ N ] ] ] [ TR[-wh] ] ] ] ] ] ]
} \\
\bottomrule
\end{tabularx}
\caption{Examples used in pre-pretraining.}
\label{tab:example_ppt}
\end{table}

\section{Calculation Examples of Learning Efficiency Metrics}
\label{appendix:efficiency_metrics}

We present an actual calculation example using the results from one trial (Seed=0) in our experiment.
When the reference point is set to $y_1 = 25{,}000$, the loss of the baseline (Non-PPT) was approximately $3.633$. 
The proposed method (\textsc{MP-Struct}) first reached this loss at $y_2 \approx 15{,}755$ steps. 
Since the number of formal language training steps is $x = 500$, the metrics are calculated as follows:

\begin{equation}
\begin{split}
    \text{MRS} &= \frac{25{,}000 - 15{,}755}{500} \\
    &= \frac{9{,}245}{500} = 18.49
\end{split}
\end{equation}

\begin{equation}
\begin{split}
    \text{Efficiency Gain} &= 1 - \frac{15{,}755 + 500}{25{,}000} \\
    &= 1 - 0.65 = 0.35
\end{split}
\end{equation}

\section{Jabberwocky Dataset Construction}
\label{appendix:jabberwocky_construction}

To evaluate the structural robustness of the models, we constructed a Jabberwocky (JW) variant of the C4 dataset.
This process aims to eliminate semantic correlations from lexical co-occurrence while strictly preserving the syntactic structure and morphological consistency of the original text.

\paragraph{Implementation Details}
We implemented the generation pipeline using the \texttt{spaCy} library with the \texttt{en\_core\_web\_sm} model.
The transformation process operates as follows:

\begin{itemize}
    \item \textbf{Fine-grained POS Tagging:} 
    We first tokenize the input text and assign fine-grained Part-of-Speech (POS) tags. 
    Unlike coarse tags (e.g., \texttt{NOUN}), fine-grained tags (e.g., \texttt{NNS} for plural nouns, \texttt{VBD} for past tense verbs) allow us to distinguish morphological forms strictly.

    \item \textbf{Content Word Identification:} 
    We identify content words defined as tokens belonging to the set of coarse categories: \texttt{\{NOUN, VERB, ADJ, ADV\}}. 
    Function words (e.g., determiners, prepositions) and punctuation are preserved to maintain the grammatical structures.

    \item \textbf{Tag-wise Shuffling (Document Level):} 
    To preserve morphological agreement (e.g., subject-verb number agreement), we strictly shuffle words within the same fine-grained tag category. 
    Specifically, we group all content words within a processing batch by their fine-grained tags and shuffle these buckets randomly.
    
    \item \textbf{Lexical Replacement with Casing Constraints:} 
    Each content word in the original sequence is replaced by a different word drawn from the corresponding shuffled tag bucket. 
    Crucially, we apply casing constraints: if the original token was capitalized (e.g., sentence initial), the replaced token is capitalized to maintain sentence boundaries.
\end{itemize}

By using fine-grained tags rather than coarse categories, our method ensures that, for instance, a singular noun is always replaced by another singular noun, and a past-tense verb by another past-tense verb. 
This guarantees that the resulting sequences preserve syntactic well-formedness despite the removal of semantic information.

\section{Impossible Language Datasets Construction}
\label{appendix:impossible_languages}

To evaluate whether the model's inductive bias aligns with human-like linguistic constraints, we constructed three ``Impossible Language'' datasets.
We adapted the perturbation logic from the official implementation of \citet{kallini-etal-2024-mission}\footnote{\url{https://github.com/jkallini/mission-impossible-language-models/}} and integrated it into our preprocessing pipeline.

While the Jabberwocky dataset operates at the document level to maintain vocabulary pools, the following transformations were applied at the \textbf{sentence level} after segmentation using \texttt{spaCy}.
We generated the datasets using the following configurations:

\begin{itemize}
    \item \textbf{SHUFFLE:} \\
    Generated with the argument \texttt{shuffle\_deterministic21}. 
    This transformation permutes tokens deterministically within a fixed local window of size $s=21$.
    By destroying local word order (n-grams) while preserving global bag-of-words statistics, this condition tests the model's reliance on local syntactic constituency.

    \item \textbf{REVERSE:} \\
    Generated with the argument \texttt{reverse\_full}.
    This operation reverses the token order of the entire sentence string ($w_1, w_2, \dots, w_n \to w_n, \dots, w_2, w_1$).
    While computationally deterministic (requiring a stack), this transformation violates the incremental, left-to-right processing constraint fundamental to human language performance.

    \item \textbf{HOP:} \\
    Generated with the argument \texttt{hop\_words4}.
    This transformation introduces a dependency based on linear counting rather than structural configuration. 
    Specifically, a functional marker is placed at a fixed linear distance of $k=4$ words after its associated verb.
    This mimics ``impossible'' grammatical rules that rely on counting word positions in the linear string, violating the structure-dependence principle of Universal Grammar.
\end{itemize}

All transformations were applied to the same subset of the C4 training data as the other conditions, ensuring comparable data volume and lexical coverage.

\section{Complexity Calibration of Abstract Conditions}
\label{appendix:complexity_calibration}
 
In \S\ref{sec:analysis_mechanism}, we parameterize our abstract formal languages by two values: $k_{\text{struct}}$, the number of bracket types used for hierarchical structure, and $k_{\text{dep}}$, the number of distinct dependency types.
We set $k_{\text{struct}}=1$ (corresponding to 1-Dyck) and $k_{\text{dep}}=4$ (corresponding to 4-Shuffle Dyck).
This choice is not arbitrary but is derived from an analysis of the dependency types inherent in the full \textsc{MP-Struct} generator.
\textsc{MP-Struct} generates sequences based on five distinct structural operations:
 
\begin{enumerate}
    \item \textbf{Structure (\textsc{Merge}):} The fundamental recursive skeleton formed by brackets (e.g., $[ \dots ]$). This corresponds to the \textbf{1-Dyck} component.
    \item \textbf{Dependencies (\textsc{Move}/\textsc{Agree}/\textsc{Select}):} Within this skeleton, four distinct types of long-distance dependencies are established. These correspond to the \textbf{4-Shuffle Dyck} component ($k=4$):
    \begin{itemize}
        \item \textbf{Type 1: Agreement (Plural).} The dependency between $T_{pl}$ and $DP_{pl}$.
        \item \textbf{Type 2: Agreement (Singular).} The dependency between $T_{sg}$ and $DP_{sg}$.
        \item \textbf{Type 3: Movement.} The dependency between a functional head (e.g., $C$) and a trace ($TR$) formed by Wh-movement.
        \item \textbf{Type 4: Selection.} The local dependency where a determiner ($D$) selects a noun ($N$).
    \end{itemize}
\end{enumerate}
 
By setting $k_{\text{struct}}=1$ and $k_{\text{dep}}=4$, we ensure that both \textbf{Generic $k$-SD} and \textbf{\textsc{MP-Struct Core}} possess the same "vocabulary size" of dependency types as the original model, allowing us to isolate the effect of topological arrangement and landmarks without confounding factors related to task complexity.

\section{\textsc{MP-Struct Core}: Design Details}
\label{appendix:core_generation}
 
The design of \textsc{MP-Struct Core} and its motivation are described in \S\ref{subsec:decomposition}.
Here we provide supplementary detail on the mapping from \textsc{MP-Struct} to its abstract counterpart.
 
\paragraph{Fixed Topology}
The derivation in \textsc{MP-Struct} follows a recursive path ($CP \to TP \to vP$).
\textsc{MP-Struct Core} strictly enforces this same hierarchical subordination, ensuring that every dependency is generated within its designated structural domain.
 
\paragraph{Abstract Landmarks}
The functional heads of \textsc{MP-Struct} are replaced by distinct abstract tokens with the following correspondence:
\begin{itemize}
    \item \textbf{Complementizer ($C$) $\to H_{CP}$:} Marks the clause boundary and movement landing site.
    \item \textbf{Tense ($T$) $\to H_{TP}$:} Marks the inflectional domain and agreement trigger.
    \item \textbf{Verb ($v$) $\to H_{VP}$:} Marks the thematic domain (argument structure).
\end{itemize}
Argument structure is fixed to a transitive frame, ensuring that $H_{CP}$, $H_{TP}$, and $H_{VP}$ serve as unambiguous, consistent landmarks across all generated sequences.

\end{document}